\documentclass{article} 
\usepackage{iclr2027_conference}
\usepackage{times}

\usepackage{amsmath,amsfonts,bm}

\def\eqref#1{equation~\ref{#1}}

\def\1{\bm{1}}

\DeclareMathAlphabet{\mathsfit}{\encodingdefault}{\sfdefault}{m}{sl}
\SetMathAlphabet{\mathsfit}{bold}{\encodingdefault}{\sfdefault}{bx}{n}

\usepackage{hyperref}
\usepackage{url}
\usepackage{graphicx}
\usepackage{booktabs}
\usepackage{pgfplots}
\pgfplotsset{compat=1.18}

\title{Dead Weights, Live Signals: Feedforward Graphs of Frozen Language Models}

\author{
  Marcus Armstrong \quad Navid Ayoobi \quad Arjun Mukherjee \\
  Department of Computer Science \\
  University of Houston \\
  Houston, TX 77204 \\
  \texttt{\{miarmstr, nyoobi\}@cougarnet.uh.edu, amukher6@central.uh.edu}
}

\iclrfinalcopy
\begin{document}

\maketitle

\begin{abstract}
We present a feedforward graph architecture in which heterogeneous
frozen large language models serve as computational nodes,
communicating through a shared continuous latent space via learned
linear projections. Building on recent work demonstrating geometric
compatibility between independently trained LLM latent
spaces~\cite{armstrong2026thinking}, we extend this finding from
static two-model steering to end-to-end trainable multi-node graphs,
where projection matrices are optimized jointly via backpropagation
through residual stream injection hooks. Three small frozen models
(Llama-3.2-1B, Qwen2.5-1.5B, Gemma-2-2B) encode the input into a
shared latent space whose aggregate signal is injected into two larger
frozen models (Phi-3-mini, Mistral-7B), whose representations feed a
lightweight cross-attention output node. With only 17.6M trainable
parameters against approximately 12B frozen, the architecture achieves
87.3\% on ARC-Challenge, 82.8\% on OpenBookQA, and 67.2\% on MMLU,
outperforming the best single constituent model by 11.4, 6.2, and 1.2
points and outperforming majority vote of the same five models by 8.2,
8.0, and 10.4 points --- benchmarks where output-level ensembling
actively degrades performance. Gradient flow through frozen model
boundaries is empirically verified to be tractable, and the output
node develops selective routing toward Phi-3-mini without supervision,
validated by a 7.8pp performance collapse upon removal.
\end{abstract}

\section{Introduction}

The prevailing paradigm for improving language model performance is
scaling: larger models, more data, more compute. Yet smaller models
trained on specialized corpora routinely outperform larger general
models on narrow tasks~\cite{phi3,qwen25}, and models trained with
different objectives or architectures exhibit systematically different
strengths at equivalent parameter counts. The information needed to
solve a given problem may already exist, distributed unevenly across
a population of existing models --- the challenge is aggregating it.

Current ensemble and multi-agent approaches pursue this at the output
level: routing queries to specialized models~\cite{mixtral}, ensembling
token distributions~\cite{ensemble_lm}, or having models critique each
other's responses~\cite{autogen,crewai}. These methods treat each model
as a black box, and their limitations are concrete: majority vote over
five heterogeneous models actually \textit{degrades} performance on
MMLU and OpenBookQA relative to the best single model, because weaker
models corrupt the vote. The geometric structures in LLM latent spaces
that encode semantic relationships, reasoning patterns, and factual
knowledge remain inaccessible to output-level approaches.

\citet{armstrong2026thinking} showed this need not be the case:
independently trained LLMs converge to geometrically compatible latent
spaces, and a linear projection suffices to translate activations
between heterogeneous architectures. We exploit this compatibility in a
new direction --- not for inference-time correction between two models,
but as a \emph{differentiable communication medium} for a feedforward
graph of frozen LLMs. Learned projection matrices translate each node's
hidden state into a shared latent space; the aggregate signal is
injected into downstream nodes via residual stream hooks; and the
entire system is trained end-to-end via backpropagation through the
frozen model boundaries. Only the projection matrices and a lightweight
cross-attention output node are updated --- approximately 17.6M
parameters against 12B frozen.

We make five contributions:

\begin{enumerate}
    \item \textbf{Tractable gradient flow.} Gradients flow cleanly
    through multiple frozen LLM boundaries, with layer-1 projection
    matrices receiving approximately 13\% of the gradient signal at
    the output node --- attenuated but sufficient for learning, with
    skip connections providing no additional benefit.
 
    \item \textbf{Strong benchmark performance.} The graph achieves
    87.3\% on ARC-Challenge, 82.8\% on OpenBookQA, and 67.2\% on
    MMLU, outperforming the best single constituent model by 11.4,
    6.2, and 1.2 percentage points respectively, and outperforming
    majority vote of the same five models by 8.2, 8.0, and 10.4
    points --- benchmarks where output-level ensembling is actively
    harmful.
 
    \item \textbf{Comprehensive ablations.} The graph outperforms
    parameter-matched learned classifiers by 9.1, 5.2, and 6.7 points
    across three benchmarks. Systematic ablations confirm robustness
    across a 10$\times$ range of injection coefficient and 8$\times$
    range of shared dimension, and a prompt ablation establishes that
    performance is stable within 1pp across all framing strategies ---
    representational diversity derives from model heterogeneity, not
    prompt engineering.
 
    \item \textbf{Emergent selective routing.} Without supervision,
    the output node develops asymmetric attention over the two
    layer-2 nodes, preferentially weighting Phi-3-mini --- a pattern
    consistent with its synthetic training corpus inducing geometrically
    regular representations, and validated by a 7.8pp performance
    collapse when Phi-3-mini is removed.
\end{enumerate}

\section{Related Work}
\label{sec:related}

\paragraph{Activation steering and residual stream intervention.}
Prior work has shown that transformer behavior can be controlled by
directly manipulating internal activations without modifying weights.
\citet{turner2024steering} demonstrated that fixed bias vectors reliably
shift model behavior across behavioral dimensions, and
\citet{zou2023representation} showed that interpretable concept directions
generalize across tasks and model families. These methods establish the
residual stream as a writable medium --- a foundational assumption of
our injection mechanism. \citet{armstrong2026thinking} extended this to
cross-architecture settings, showing that a linear projection suffices
to translate activation vectors between independently trained LLMs and
that injecting translated representations corrects model behavior without
weight updates. Three of their findings directly inform our design: the
affine compatibility of cross-architecture latent spaces (justifying
linear projection matrices), the superiority of dense over sparse
projections ($R^2 \approx 0.50$ for Ridge vs.\ $\approx 0.01$ for
Lasso), and domain orthogonality of offline-fitted projections
(motivating end-to-end training against a task loss). Our work
generalizes theirs from static two-model steering to end-to-end
trainable multi-node graphs, transforming geometric compatibility from
an analytical tool into a differentiable communication medium.

\paragraph{Model composition and ensemble methods.}
Mixture-of-experts architectures~\cite{mixtral} route tokens to
specialized subnetworks within a single model, while multi-agent
frameworks~\cite{autogen,crewai} coordinate LLMs through natural
language. Both operate at the token level; internal representations
remain inaccessible across model boundaries. Model
stitching~\cite{bansal2021revisiting} demonstrated that independently
trained networks can be composed via a learned linear adapter,
establishing affine compatibility as an empirically grounded prior. We
extend this from pairwise adapters to a full graph topology with
jointly optimized projection matrices. The Platonic Representation
Hypothesis~\cite{platonic} provides theoretical grounding: neural
networks converge to a shared underlying representation of reality,
with architecture-specific differences reducible to coordinate
transformations.

\paragraph{Frozen backbone adaptation.}
Parameter-efficient methods such as LoRA~\cite{lora} and prefix
tuning~\cite{prefix_tuning} adapt a single frozen model with minimal
parameters, but do not address how multiple frozen models might
communicate. \citet{frozen_gradient_rectifier} showed that frozen LLM
blocks act as gradient coherence rectifiers, producing stable gradient
flow through frozen boundaries --- directly addressing the theoretical
concern about our architecture's trainability. \citet{llm_autodiff}
studied multi-node LLM computation graphs and recommended skip
connections and auxiliary losses to mitigate gradient attenuation; our
empirical analysis finds neither necessary at two-layer depth
(Section~\ref{sec:analysis}). Our work differs from all of the above
by composing \emph{multiple} frozen models into a graph where
inter-node communication occurs in a shared continuous latent space
learned from the task signal.

\section{Method}
\label{sec:method}

We define a feedforward graph $\mathcal{G} = (\mathcal{V}, \mathcal{E})$
where each node $v \in \mathcal{V}$ is a frozen pretrained LLM and each
directed edge $(u, v) \in \mathcal{E}$ carries a learned projection
matrix $W_{uv} \in \mathbb{R}^{d_s \times d_u}$. Only the projection
matrices and a lightweight output node are updated during training; all
LLM weights $\theta_v$ remain frozen throughout.

\subsection{Architecture}

\paragraph{Layer 1: Diverse Encoding.}
Three heterogeneous frozen LLMs --- Llama-3.2-1B ($d{=}2048$),
Qwen2.5-1.5B ($d{=}1536$), Gemma-2-2B ($d{=}2304$) --- each receive
the input with a distinct framing prefix (``analyze the factual
content'', ``analyze the reasoning structure'', ``analyze the language
and framing''). Each node produces a hidden state at relative depth
$l_T = 0.90$, extracted at the final token position and L2-normalized:
\begin{equation}
    \mathbf{h}_i = \frac{\text{hidden}(x \oplus p_i,\, \theta_i,\, l_T)}
    {\|\text{hidden}(x \oplus p_i,\, \theta_i,\, l_T)\|_2}
    \quad i \in \{1,2,3\}
\end{equation}
We extract at depth 0.90 because deeper layers yield more fully resolved
semantic representations; L2 normalization ensures the projection learns
directional alignment rather than magnitude differences. As established
empirically in Section~\ref{sec:experiments}, the specific framing
content is non-critical, performance varies by less than 1pp across
all prompt strategies including identical prompts for all three nodes;
representational diversity derives from the architectural heterogeneity
of the frozen models rather than from the prefix choice.

\paragraph{Shared Latent Space.}
Each hidden state is projected and averaged into a shared vector:
\begin{equation}
    \mathbf{z}_1 = \tfrac{1}{3}\sum_{i=1}^{3} W_i\,\mathbf{h}_i
    \qquad W_i \in \mathbb{R}^{d_s \times d_i},\quad d_s = 1024
\end{equation}
$d_s = 1024$ is deliberately neutral --- not equal to any constituent
model's hidden dimension --- so every projection must perform genuine
geometric work rather than approximate an identity mapping.

\paragraph{Layer 2: Injection and Refinement.}
$\mathbf{z}_1$ is injected into the residual stream of each layer-2
node --- Phi-3-mini ($d{=}3072$) and Mistral-7B ($d{=}4096$) --- at
relative depth $l_S = 0.75$. Since $d_s = 1024 \ne d_j$, we linearly
interpolate $\mathbf{z}_1$ to the target dimension before injection;
this interpolation is differentiable, preserving gradient flow to the
layer-1 matrices. The blend follows the magnitude-rescaling formulation
of \citet{armstrong2026thinking}:
\begin{equation}
    \mathbf{h}^\prime = (1-\alpha)\,\mathbf{h}_\text{node}
    + \alpha\!\left(\tilde{\mathbf{z}}_1 \cdot
    \|\mathbf{h}_\text{node}\|_2\right),
    \quad \alpha = 0.25
\end{equation}
where $\tilde{\mathbf{z}}_1$ is the interpolated injection vector.
Magnitude rescaling ensures the injected signal is compatible with
the residual stream regardless of dimensional differences. We use $\alpha = 0.25$; our ablation confirms representations become
fragile at higher values ($\alpha = 0.75$ collapses to 24.4\% on
ARC-Challenge, Section~\ref{sec:experiments}), consistent with prior
work on injection fragility~\cite{armstrong2026thinking}. Each layer-2 node completes its forward
pass and its hidden state is projected:
\begin{equation}
    \mathbf{z}_2^{(j)} = W_{j+3}\,\mathbf{h}_j^\prime
    \quad j \in \{4,5\}
\end{equation}

\begin{figure}[t]
\centering
\includegraphics[width=0.82\columnwidth]{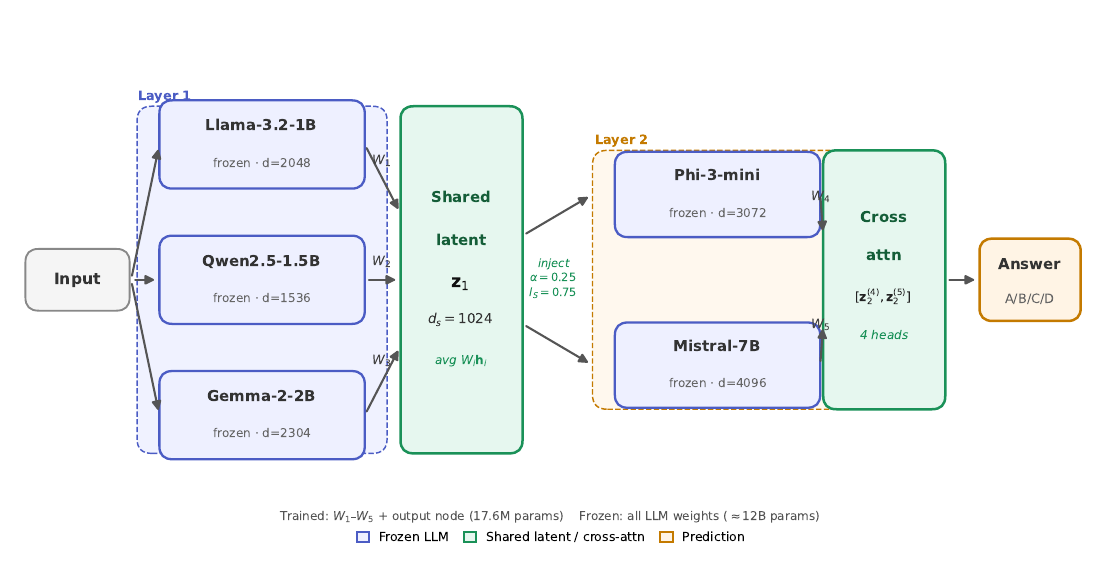}
\caption{
    \textbf{Frozen LLM Graph Architecture.} Three heterogeneous frozen
    layer-1 models encode the input with distinct framings; their hidden
    states are projected via $W_1, W_2, W_3$ into a shared latent space
    and averaged to form $\mathbf{z}_1$, which is injected into two
    frozen layer-2 models at $l_S{=}0.75$. Layer-2 representations are
    projected via $W_4, W_5$ into a second shared space, over which a
    cross-attention output node produces the final prediction. Only the
    five projection matrices and output node (17.6M params) are trained.
}
\label{fig:architecture}
\end{figure}

\paragraph{Output Node.}
A cross-attention output node aggregates both layer-2 representations:
\begin{equation}
    \mathbf{o} = \text{LayerNorm}\!\left(
        \text{MHA}\!\left(\mathbf{q},\,
        [\mathbf{z}_2^{(4)}, \mathbf{z}_2^{(5)}],\,
        [\mathbf{z}_2^{(4)}, \mathbf{z}_2^{(5)}]\right)\right),
    \quad \hat{y} = \text{softmax}(W_\text{cls}\,\mathbf{o})
\end{equation}
where $\mathbf{q} \in \mathbb{R}^{d_s}$ is a learned query, MHA uses
4 heads, and $W_\text{cls} \in \mathbb{R}^{4 \times d_s}$ maps to the
four answer choices. The attention weights implicitly route the
prediction between the two layer-2 nodes based on their representations'
compatibility with the learned query. Figure~\ref{fig:architecture}
illustrates the complete architecture.

\subsection{Training}

We minimize cross-entropy loss $\mathcal{L} = -\log\hat{y}_{c^\star}$
where $c^\star$ is the correct answer index. Gradients flow backward
through the output node, through $W_4$ and $W_5$, through the
differentiable injection operations, and into $W_1$, $W_2$, $W_3$.
Frozen LLM weights receive no gradients.

We use AdamW~\cite{adamw} with $\eta_W = 10^{-3}$ for projection
matrices and $\eta_\text{out} = 10^{-4}$ for the output node, weight
decay $10^{-4}$, gradient clip 1.0, effective batch size 8, and cosine
annealing. All models are loaded in \texttt{bfloat16} via
HuggingFace~\cite{wolf2020transformers}; the full graph fits on a
single 80GB A100. Injection hooks are registered immediately before
each forward pass and removed afterward to prevent cross-example
contamination. Full hyperparameters are in Table~\ref{tab:hparams};
trainable parameter counts are in Table~\ref{tab:params} (Appendix).

\section{Experiments}
\label{sec:experiments}

\subsection{Datasets and Baselines}

We evaluate on three multiple-choice benchmarks covering distinct
reasoning demands. \textbf{MMLU}~\cite{mmlu} provides 14,042 training
and 1,531 validation questions across 57 subjects (STEM, humanities,
social sciences). \textbf{ARC-Challenge}~\cite{arc} provides 1,119
training and 1,165 test questions drawn from standardized science
tests. \textbf{OpenBookQA}~\cite{openbookqa} provides 4,957 training
and 500 test questions requiring integration of a core science fact
with commonsense inference. Together the three benchmarks cover broad
factual recall, standardized scientific reasoning, and fact-integrated
inference.

We compare against three baselines under the same zero-shot greedy
decoding protocol. \textbf{Single-model greedy}: each constituent
model evaluated individually with no learned components.
\textbf{Majority vote}: output-level ensemble over all five constituent
models with no learned parameters, representing the strongest
parameter-free combination baseline. \textbf{Parameter-matched learned
head}: an MLP classifier trained on the frozen hidden states of the
strongest single-model baseline, with parameter budget matched to the
graph network's 17.6M parameters (16.8M for Mistral-7B on MMLU;
22.8M for Qwen2.5-1.5B on ARC and OpenBookQA). This baseline isolates
whether gains arise from the communication mechanism or from the learned
classifier alone.

\subsection{Results}

\begin{table}[h]
\centering
\caption{
    Results on MMLU, ARC-Challenge (ARC), and OpenBookQA (OBQA).
    Best single-model result per benchmark is underlined.
    Majority vote requires no learned parameters.
    The learned head uses Mistral-7B for MMLU and Qwen2.5-1.5B
    for ARC and OBQA.
}
\label{tab:main}
\begin{tabular}{lccc}
\toprule
\textbf{Method} & \textbf{MMLU} & \textbf{ARC} & \textbf{OBQA} \\
\midrule
\multicolumn{4}{l}{\textit{Single-model greedy}} \\
\quad Llama-3.2-1B-Instruct    & 44.0\% & 50.8\% & 56.6\% \\
\quad Qwen2.5-1.5B-Instruct    & 59.0\% & \underline{75.9\%} & \underline{76.6\%} \\
\quad Gemma-2-2B-IT             & 60.0\% & 67.1\% & 60.8\% \\
\quad Phi-3-mini-4K-Instruct    & 43.0\% & 75.3\% & 62.2\% \\
\quad Mistral-7B-Instruct-v0.3  & \underline{66.0\%} & 73.9\% & 70.4\% \\
\midrule
\multicolumn{4}{l}{\textit{Output-level ensemble (no learned parameters)}} \\
\quad Majority vote (5 models) & 56.8\% & 79.1\% & 74.8\% \\
\midrule
\multicolumn{4}{l}{\textit{Parameter-matched learned head on best single model}} \\
\quad Head on Mistral-7B (16.8M)    & 60.5\% & ---    & ---    \\
\quad Head on Qwen2.5-1.5B (22.8M) & ---    & 78.2\% & 77.6\% \\
\midrule
\multicolumn{4}{l}{\textit{Ours}} \\
\quad \textbf{Frozen LLM graph (17.6M)} & \textbf{67.2\%} & \textbf{87.3\%} & \textbf{82.8\%} \\
\midrule
\multicolumn{4}{l}{\textit{Margins}} \\
\quad vs.\ best single model & +1.2pp & +11.4pp & +6.2pp \\
\quad vs.\ majority vote     & +10.4pp & +8.2pp  & +8.0pp \\
\quad vs.\ learned head      & +6.7pp & +9.1pp  & +5.2pp \\
\bottomrule
\end{tabular}
\end{table}

The graph network outperforms all baselines on every benchmark.
The majority vote results are particularly revealing: while majority
vote over five heterogeneous models improves over the best single
model on ARC-Challenge (+3.2pp), it \textit{degrades} performance on
MMLU ($-$9.2pp) and OpenBookQA ($-$1.8pp), where the weaker constituent
models (Llama-3.2-1B at 44\%, Phi-3-mini at 43\%) corrupt the vote.
The graph network achieves +10.4pp over majority vote on MMLU and
+8.0pp on OpenBookQA, benchmarks where output-level ensembling is
actively harmful, demonstrating that latent-space communication
extracts useful representational signal from weaker models without
inheriting their incorrect output predictions. This is qualitatively
different from output-level combination, not merely quantitatively
better.
 
The margins against the parameter-matched learned head (6.7--9.1pp)
confirm that the gains arise from the communication mechanism rather
than the classifier. A notable secondary finding is that the learned
head on Mistral-7B \textit{underperforms} Mistral's greedy decoding
on MMLU (60.5\% vs.\ 66.0\%), consistent with Mistral's instruction
tuning optimizing its output distribution in a way that is not
recoverable from the penultimate representation alone. The graph
network circumvents this by using Mistral as a refinement node rather
than a representation to classify directly.

\subsection{Error Analysis}

To characterize the gains mechanistically, we compare graph and
individual model predictions question-by-question on the 1,165
ARC-Challenge test examples. Of these, 54.4\% involve partial
inter-model disagreement, and on these the graph consistently resolves
disagreement toward the correct answer, the primary source of the
performance gain. Ten questions (0.9\%) are recovered by the graph
despite all five individual models failing, while 95 regressions
(8.2\%) represent a real cost of injection. The majority vote
comparisons sharpen the mechanistic interpretation: on ARC, where
model quality is uniform, majority vote improves over the best single
model (+3.2pp) yet the graph exceeds it by a further +8.2pp. On MMLU
and OpenBookQA, majority vote actively degrades performance ($-$9.2pp
and $-$1.8pp), while the graph improves on both --- confirming that
latent-space aggregation extracts useful signal from weaker models
without inheriting their incorrect outputs.
 
\subsection{Ablation Studies}
 
We ablate the three primary design choices of the architecture on
ARC-Challenge using 3-epoch runs (full 4-epoch results reported above).
All ablations use fixed random seed 42.
 
\begin{table}[h]
\centering
\caption{
    Ablation results on ARC-Challenge test set (3-epoch runs).
    Full architecture achieves 87.7\% at 3 epochs vs.\ 87.3\% at 4
    epochs, consistent with run-to-run variance. $\Delta$ is relative
    to the full architecture within each experiment group.
    $\star$ denotes the reported configuration.
}
\label{tab:ablations}
\begin{tabular}{llcc}
\toprule
\textbf{Experiment} & \textbf{Variant} & \textbf{Test} & $\boldsymbol{\Delta}$ \\
\midrule
\multicolumn{4}{l}{\textit{(A) Injection coefficient $\alpha$}} \\
\quad & 0.10           & 86.0\% & $-$1.7pp \\
\quad & 0.25$^\star$   & 86.6\% & --- \\
\quad & 0.50           & 87.0\% & $+$0.4pp \\
\quad & 0.75           & 24.4\% & $-$62.2pp \\
\midrule
\multicolumn{4}{l}{\textit{(B) Shared latent dimension $d_s$}} \\
\quad & 256            & 87.3\% & $-$0.6pp \\
\quad & 1024$^\star$   & 87.9\% & --- \\
\quad & 2048           & 86.4\% & $-$1.5pp \\
\midrule
\multicolumn{4}{l}{\textit{(C) Node removal}} \\
\quad & Full$^\star$         & 87.7\% & --- \\
\quad & Drop Llama-3.2-1B   & 86.8\% & $-$0.9pp \\
\quad & Drop Qwen2.5-1.5B   & 86.1\% & $-$1.6pp \\
\quad & Drop Gemma-2-2B     & 86.5\% & $-$1.2pp \\
\quad & Drop Phi-3-mini     & 79.9\% & $-$7.8pp \\
\quad & Drop Mistral-7B     & 87.2\% & $-$0.5pp \\
\quad & 2$\times$L1 only    & 86.7\% & $-$1.0pp \\
\midrule
\multicolumn{4}{l}{\textit{(D) L1 prompt strategy}} \\
\quad & Analytical$^\star$  & 87.8\% & --- \\
\quad & Distinct neutral    & 87.3\% & $-$0.5pp \\
\quad & Identical neutral   & 87.2\% & $-$0.6pp \\
\quad & No prefix           & 86.8\% & $-$1.0pp \\
\bottomrule
\end{tabular}
\end{table}
 
\paragraph{Injection coefficient (A).}
Performance is stable within 1.7pp for $\alpha \in \{0.10, 0.25,
0.50\}$; at $\alpha = 0.75$ it collapses to near-random (24.4\%),
directly replicating the representation fragility reported
by~\citet{armstrong2026thinking}.

\paragraph{Shared dimension (B).}
Performance is robust across an 8$\times$ range. $d_s = 256$ with
4$\times$ fewer projection parameters achieves $-$0.6pp, suggesting
the shared space develops geometrically structured representations
rather than requiring high capacity.

\paragraph{Node removal (C).}
Dropping Phi-3-mini causes a 7.8pp collapse, the functional
validation of the routing finding. All three L1 nodes contribute
comparably (0.9--1.6pp when dropped), confirming the averaging
operation does not degenerate. Dropping Mistral-7B costs only 0.5pp,
consistent with its weaker gradient signal throughout training.

\paragraph{Prompt strategy (D).}
Performance varies by less than 1pp across all strategies, including
identical prompts for all three L1 nodes. Representational diversity
derives from model heterogeneity, not prompt engineering.

\subsection{Training Dynamics}

\begin{figure}[t]
\centering
\includegraphics[width=\columnwidth]{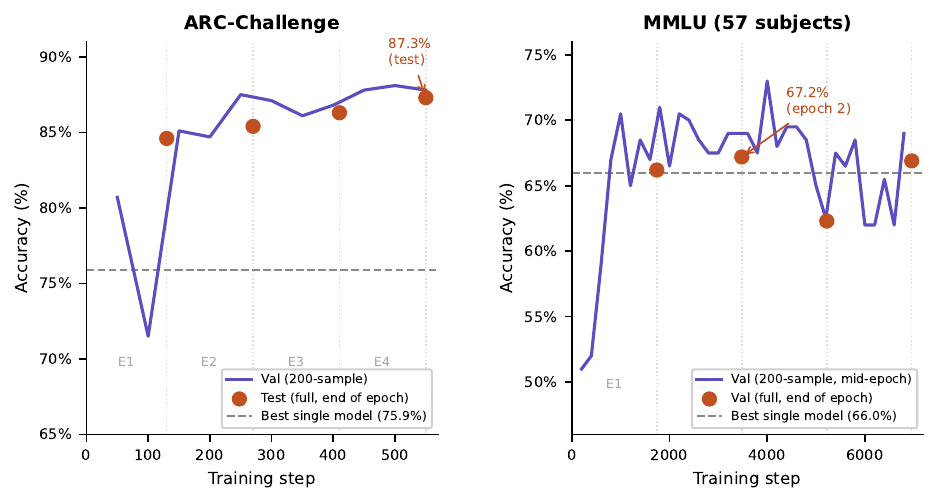}
\caption{
    Validation accuracy over training on ARC-Challenge (left) and
    MMLU (right). Violet line: mid-epoch accuracy on 200-example
    subset. Coral markers: end-of-epoch full evaluation. Dashed line:
    best single-model baseline. ARC-Challenge reaches 80.7\% after
    50 steps; MMLU peaks at 67.2\% at epoch~2.
}
\label{fig:training}
\end{figure}

Figure~\ref{fig:training} shows validation accuracy over training.
ARC-Challenge converges rapidly --- 80.7\% after just 50 gradient
steps --- reflecting the immediate utility of aggregated multi-node
representations. OpenBookQA peaks at 85.4\% validation mid-epoch~2
and plateaus; training stops at epoch~3. MMLU converges more slowly
over 14,042 training examples, peaking at 67.2\% at epoch~2 under
cosine annealing. Full hyperparameters are in Table~\ref{tab:hparams}.

\begin{table}[h]
\centering
\caption{Training hyperparameters for all graph network experiments.}
\label{tab:hparams}
\begin{tabular}{ll}
\toprule
\textbf{Hyperparameter} & \textbf{Value} \\
\midrule
Optimizer                         & AdamW \\
LR ($W_1$--$W_5$) / output node   & $10^{-3}$ / $10^{-4}$ \\
Weight decay                      & $10^{-4}$ \\
Effective batch size              & 8 \\
LR schedule                       & Cosine annealing \\
Gradient clip                     & 1.0 \\
Epochs                            & 4 (MMLU, ARC); 3 (OBQA) \\
$l_S$ / $l_T$ / $\alpha$          & 0.75 / 0.90 / 0.25 \\
$d_s$ / output heads              & 1,024 / 4 \\
Precision / hardware              & bfloat16 / A100 80GB \\
\bottomrule
\end{tabular}
\end{table}

\section{Analysis}
\label{sec:analysis}
 
\subsection{Gradient Flow Through Frozen Model Boundaries}
 
We validate gradient flow with a minimal two-node graph: Llama-3.2-1B
as source, Qwen2.5-1.5B as destination, connected by a single
projection matrix $W \in \mathbb{R}^{1536 \times 2048}$ and a
four-class output head. We measure gradient norms, the $W$-to-head
gradient ratio, and a permutation control (target activations randomly
shuffled to destroy semantic correspondence while preserving marginal
statistics).
 
Results are summarized in Table~\ref{tab:gradient}. Gradients flow
cleanly: $W$ receives approximately 13\% of the output head's gradient
signal --- attenuated but well above catastrophic collapse, and
comparable to attenuation across equivalent depth in standard
feedforward networks. The permutation $R^2$ of $-0.243$ versus Ridge
$R^2$ of $0.299$ confirms the projection captures genuine semantic
geometry rather than distributional artifacts. Skip connections produce
a 1.00$\times$ improvement, confirming they are unnecessary at this
depth and simplifying the final architecture.
 
\begin{table}[h]
\centering
\caption{Gradient flow validation (Llama-1B $\to$ Qwen-1.5B). Negative
permutation $R^2$ confirms genuine cross-architecture alignment.}
\label{tab:gradient}
\begin{tabular}{lc}
\toprule
\textbf{Metric} & \textbf{Value} \\
\midrule
Ridge projection $R^2$       & 0.299 \\
Permutation control $R^2$    & $-$0.243 \\
$W$ gradient max             & $2.48 \times 10^{-1}$ \\
$W$ gradient mean            & $1.81 \times 10^{-3}$ \\
Grad norm ratio ($W$ / head) & 0.130 \\
Skip connection improvement  & 1.00$\times$ (neutral) \\
\bottomrule
\end{tabular}
\end{table}
 
\subsection{Projection Matrix Gradient Norms During Training}
 
\begin{figure}[t]
\centering
\includegraphics[width=\columnwidth]{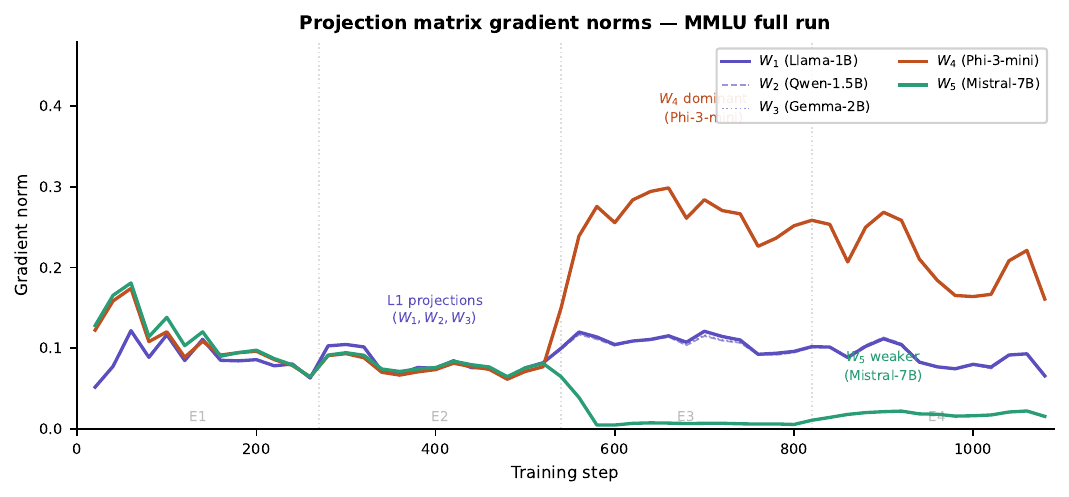}
\caption{
    Projection matrix gradient norms over training (MMLU, 57 subjects).
    $W_1$--$W_3$ (violet) converge to nearly identical norms, showing
    layer-1 matrices do not specialize. $W_4$ (Phi-3-mini, coral)
    consistently dominates $W_5$ (Mistral-7B, teal), reflecting
    emergent selective routing. All five matrices remain active.
}
\label{fig:gradnorms}
\end{figure}

Figure~\ref{fig:gradnorms} shows gradient norms for $W_1$--$W_5$
across 4 epochs of MMLU training. Three findings emerge.
 
\textbf{No dead nodes.} All five projection matrices receive nonzero
gradients throughout training, confirming all communication channels
remain active. Heterogeneous layer-1 nodes are essential here ---
identical models would produce symmetric gradients and degenerate to
the same solution.
 
\textbf{Layer-2 projections receive stronger gradients.} $W_4$ and
$W_5$ consistently exhibit larger norms than $W_1$--$W_3$, as
expected: layer-2 matrices feed directly into the output node while
layer-1 matrices must traverse two frozen model boundaries first. The
ratio is consistent with the 0.130 measured in the validation
experiment.
 
\textbf{Layer-1 projections do not specialize.} $W_1$, $W_2$, and
$W_3$ maintain nearly identical gradient norms throughout training.
We attribute this to the averaging operation: if two matrices produce
similar projections, the third receives no gradient pressure to
diverge. The prompt ablation (Section~\ref{sec:experiments})
establishes that this non-specialization is not a failure mode ---
performance is stable within 1pp across all prompt strategies,
including identical prompts for all three L1 nodes. Representational
diversity therefore arises from the architectural and training
heterogeneity of the frozen constituent models themselves, not from
prompt differentiation or from learned specialization of the projection
matrices. The projection matrices learn to extract complementary
geometric structure from models with different inductive biases
regardless of input framing.
 
\subsection{Emergent Routing in the Output Node}
 
Without any supervision on routing, the output node develops a
consistent and substantial asymmetry between $W_4$ (Phi-3-mini) and
$W_5$ (Mistral-7B). On MMLU, $W_4$ gradient norms exceed $W_5$ by
10--30$\times$ in early training, settling to 5--10$\times$ by epoch
4. On ARC-Challenge the asymmetry is larger early and partially closes
by the final epoch.
 
This asymmetry is a direct consequence of the cross-attention weights:
a higher weight on $W_4$ means $W_4$ receives a stronger backward
signal, so the gradient ratio reflects implicit routing. We attribute
Phi-3-mini's dominant role to its synthetic training corpus, which
prior work has shown produces geometrically regular representations
more amenable to external redirection~\cite{armstrong2026thinking}.
The network discovers this property autonomously, without any prior
indicating a preference, because Phi-3-mini's representations
generate more consistent gradient signals. The node removal ablation
validates this interpretation directly: dropping Phi-3-mini causes a
7.8pp collapse on ARC-Challenge, the largest degradation of any single
node, while dropping Mistral-7B costs only 0.5pp. $W_5$ remains active
throughout training (an order of magnitude above zero), confirming
Mistral-7B contributes a secondary but nonzero signal. The
benchmark-specificity of the routing asymmetry (larger on ARC than
MMLU) is consistent with domain-dependent steerability, though
controlled ablations are needed to disentangle this from dataset size
and training duration effects.
 
\section{Conclusion}
\label{sec:conclusion}

We introduced frozen LLM graphs: heterogeneous pretrained LLMs as
computational nodes communicating through a shared continuous latent
space via end-to-end learned projections. With only 17.6M trainable
parameters against 12B frozen, the architecture achieves 87.3\% on
ARC-Challenge (+11.4pp over the best single model, +9.1pp over a
parameter-matched head, +8.2pp over majority vote of the same five
models), 82.8\% on OpenBookQA (+6.2pp, +5.2pp, +8.0pp), and 67.2\%
on MMLU (+1.2pp, +6.7pp, +10.4pp). The majority vote comparisons are
particularly informative: output-level ensembling of the same five
models actively degrades performance on MMLU and OpenBookQA relative
to the best single model, while latent-space communication consistently
improves on all three --- establishing that the gains arise from
geometric aggregation in representation space rather than from
output-level model combination. Gradient flow through frozen model boundaries is tractable at 13\%
of output-node signal strength; the output node develops asymmetric
routing toward Phi-3-mini without supervision, validated by a 7.8pp
collapse upon removal; and performance is stable within 1pp across
all prompt strategies, confirming representational diversity arises
from model heterogeneity.
 
\paragraph{Limitations.}
\textit{Single training runs}: all results use one run per benchmark;
the MMLU +1.2pp margin should be interpreted cautiously.
\textit{Scheduler instability}: cosine annealing interacts poorly with
checkpoint resumption, causing oscillating MMLU validation accuracy;
warmup-then-constant schedules should be preferred in future work.
\textit{Multiple-choice only}: generalization to open-ended generation
and longer contexts is untested, as the current output node is a
fixed-class classifier rather than a generative head.
\textit{Regression cost}: 8.2\% of ARC-Challenge test questions are
answered incorrectly by the graph despite at least one individual model
being correct, representing a real cost of residual stream injection
that selective injection strategies may mitigate.
 
\paragraph{Future directions.}
The most impactful near-term extensions are: replacing fixed average
pooling with learned attention-weighted pooling over layer-1 nodes to
enable dynamic per-query routing; scaling to deeper graphs via
multi-GPU deployment; and warm-starting projection matrices from
offline Ridge solutions to accelerate convergence. The frozen
constraint ensures that alignment properties established during
pretraining are preserved throughout composition --- as the ecosystem
of open-weight LLMs grows, treating them as composable graph nodes
rather than monolithic systems to retrain becomes increasingly
practical.

\section*{AI Use Statement}
 
The authors used Claude (Anthropic) as an AI assistant during the
preparation of this work. Specifically, Claude was used to assist with
the generation of training and evaluation code (including the graph
network forward pass, residual stream injection hooks, baseline
scripts, and diagnostic logging), and to assist with drafting and
editing prose across multiple sections of the manuscript. All
experimental results were produced and verified by the authors. All
scientific claims, architectural decisions, experimental design,
and interpretations are solely the work of the authors. No AI
tools were used to generate or fabricate experimental data.

\section*{Reproducibility Statement}
 
We have taken the following steps to facilitate reproducibility of
our results. Full training hyperparameters are reported in
Table~\ref{tab:hparams}. The trainable parameter breakdown is
provided in Appendix~\ref{app:params}. All five constituent models
are publicly available via HuggingFace Transformers~\cite{wolf2020transformers}
under their respective licenses: Llama-3.2-1B-Instruct
(\texttt{meta-llama/Llama-3.2-1B-Instruct}), Qwen2.5-1.5B-Instruct
(\texttt{Qwen/Qwen2.5-1.5B-Instruct}), Gemma-2-2B-IT
(\texttt{google/gemma-2-2b-it}), Phi-3-Mini-4K-Instruct
(\texttt{microsoft/Phi-3-mini-4k-instruct}), and
Mistral-7B-Instruct-v0.3
(\texttt{mistralai/Mistral-7B-Instruct-v0.3}). All experiments
used a fixed random seed of 42 and were conducted on a single
NVIDIA A100 80GB GPU. Training and evaluation code, including
the graph network forward pass, residual stream injection hooks,
and all baseline scripts, is provided as supplementary material.
Benchmark datasets are publicly available: MMLU via
\texttt{cais/mmlu}, ARC-Challenge via \texttt{allenai/ai2\_arc},
and OpenBookQA via \texttt{allenai/openbookqa} on the HuggingFace
Datasets hub.

\section*{Ethics Statement}
 
This work composes publicly available open-weight language models
into a graph architecture for multiple-choice question answering.
All constituent models are released under permissive research
licenses and no proprietary models or private data are used. The
work does not involve human subjects, does not collect or release
new datasets, and does not introduce capabilities beyond those of
the underlying pretrained models. The frozen constraint, under
which all LLM weights remain unchanged throughout training, 
ensures that safety and alignment properties established during
pretraining and instruction tuning are preserved in the composed
system. We identify no foreseeable negative societal impacts
specific to this work.

\bibliographystyle{abbrvnat}
\bibliography{references} 


\appendix
 
\section{Trainable Parameter Breakdown}
\label{app:params}
 
Table~\ref{tab:params} provides the full parameter count for each
trainable component of the frozen LLM graph. All five LLM backbone
weights are frozen throughout training; only the projection matrices
and output node receive gradient updates.
 
\begin{table}[h]
\centering
\caption{Trainable parameter breakdown. The five projection matrices
and output node together constitute 17,580,036 parameters ---
approximately 0.15\% of the total frozen parameter count across all
five constituent models ($\approx$12B).}
\label{tab:params}
\begin{tabular}{lrr}
\toprule
\textbf{Component} & \textbf{Dimensions} & \textbf{Parameters} \\
\midrule
$W_1$ (Llama-3.2-1B $\to$ shared)   & $1024 \times 2048$ & 2,098,176 \\
$W_2$ (Qwen2.5-1.5B $\to$ shared)   & $1024 \times 1536$ & 1,573,888 \\
$W_3$ (Gemma-2-2B $\to$ shared)     & $1024 \times 2304$ & 2,360,320 \\
$W_4$ (Phi-3-mini $\to$ shared)     & $1024 \times 3072$ & 3,146,752 \\
$W_5$ (Mistral-7B $\to$ shared)     & $1024 \times 4096$ & 4,195,328 \\
Output node (attn + classifier)      & ---                & 4,205,572 \\
\midrule
\textbf{Total trainable}             &                    & \textbf{17,580,036} \\
\midrule
\textit{Frozen (all LLM weights)}    & ---                & \textit{$\approx$12B} \\
\bottomrule
\end{tabular}
\end{table}
 
The output node's 4,205,572 parameters break down as follows: the
4-head multi-head attention over $d_s = 1024$ contributes
$4 \times (1024^2 / 4) \times 3 + 1024^2 \approx 4{,}194{,}304$
parameters for the key, query, and value projections plus output
projection, and the final linear classifier $W_\text{cls} \in
\mathbb{R}^{4 \times 1024}$ contributes 4,096 parameters plus a
learned query vector of 1,024 and LayerNorm parameters.
 
\section{Gradient Flow Validation: Full Details}
\label{app:gradient}
 
Section~\ref{sec:analysis} summarizes the gradient flow validation
experiment. We provide full experimental details here for
reproducibility.
 
\subsection{Setup}
 
We construct a minimal two-node graph consisting of
Llama-3.2-1B-Instruct (source, $d = 2048$) and
Qwen2.5-1.5B-Instruct (destination, $d = 1536$), connected by a
single projection matrix $W \in \mathbb{R}^{1536 \times 2048}$ and
a four-class linear output head $H \in \mathbb{R}^{4 \times 1536}$.
 
The forward pass proceeds as follows: (1) encode a factual question
with Llama-3.2-1B, extracting the final-token hidden state at layer
24 of 32 (depth 0.75); (2) L2-normalize the resulting vector
$\mathbf{h}_\text{src} \in \mathbb{R}^{2048}$; (3) project via $W$
to obtain $\mathbf{z} = W\mathbf{h}_\text{src} \in \mathbb{R}^{1536}$;
(4) inject $\mathbf{z}$ into Qwen's residual stream at layer 20 of 28
(depth 0.75) via a forward hook using $\alpha = 0.25$; (5) pass the
question through Qwen with the modified residual stream; (6) extract
the final-token hidden state at layer 25 of 28 (depth 0.90); (7) pass
through the output head to obtain four-class logits.
 
We initialize $W$ and $H$ with standard normal weights scaled by 0.01.
Both are registered as trainable parameters with
\texttt{requires\_grad=True}. The Llama and Qwen weights are loaded
with \texttt{requires\_grad=False}.
 
\subsection{Cross-Architecture Alignment}
 
Prior to the gradient test, we verify cross-architecture semantic
alignment using Ridge regression. We collect 50 paired forward passes
(simple factual questions) through Llama-3.2-1B and Qwen2.5-1.5B,
extract final-token hidden states at depth 0.75 from both, and fit
a Ridge regression mapping Llama representations to Qwen
representations ($\lambda = 1.0$).
 
We fit a second Ridge regression with the rows of the Qwen activation
matrix randomly permuted (permutation control), destroying semantic
correspondence while preserving marginal statistics.
 
Results: Ridge $R^2 = 0.299$; Permutation $R^2 = -0.243$. The
negative permutation $R^2$ confirms the alignment reflects genuine
cross-architecture semantic geometry rather than distributional
properties. This is lower than the $R^2 \approx 0.50$ reported by
\citet{armstrong2026thinking} for similar-scale pairs; we attribute
this to our smaller validation set (50 vs.\ 817 questions), smaller
model scales (1B/1.5B vs.\ up to 8B), and cross-family pairing
(Llama-Qwen vs.\ intra-family pairings which achieve $R^2 = 0.684$).
Critically, \citet{armstrong2026thinking} find near-zero correlation
($r \approx -0.07$) between $R^2$ and behavioral correction rate,
and our end-to-end training optimizes for task performance directly
rather than $R^2$, so the lower offline $R^2$ does not imply inferior
learned projections.
 
\subsection{Gradient Norms}
 
A single forward-backward pass through the two-node graph produces
the following results. All values are computed under a cross-entropy
loss against a dummy four-class target.
 
\begin{itemize}
    \item $W$ gradient maximum: $2.48 \times 10^{-1}$
    \item $W$ gradient mean: $1.81 \times 10^{-3}$
    \item Output head $H$ gradient maximum: $1.91 \times 10^{0}$
    \item Output head $H$ gradient mean: $1.39 \times 10^{-2}$
    \item Gradient norm ratio $\|{\nabla W}\|_F / \|{\nabla H}\|_F$: 0.130
\end{itemize}
 
The ratio of 0.130 indicates that $W$, which must receive gradients
through one frozen model boundary (Qwen), receives approximately 13\%
of the signal strength available at the output head. This is
consistent with gradient attenuation in standard feedforward networks
of equivalent depth and far above the catastrophic near-zero collapse
sometimes hypothesized for frozen transformer boundaries.
 
\subsection{Skip Connection Ablation}
 
We test whether a direct skip connection from $\mathbf{z}$ to the
output head (bypassing Qwen entirely) improves gradient flow to $W$.
With skip connection: gradient norm ratio = 0.130 (identical).
Without skip connection: gradient norm ratio = 0.130. Improvement
factor: $1.00\times$ (neutral). Skip connections provide no benefit
at two-layer depth and are omitted from the final architecture.
 
\section{Cross-Architecture Alignment Quality}
\label{app:alignment}
 
This section provides extended discussion of our offline cross-architecture
alignment quality relative to \citet{armstrong2026thinking}, which was
removed from the main paper for space.
 
Our Ridge $R^2$ of 0.299 for the Llama-1B $\to$ Qwen-1.5B pair is
lower than the $R^2 \approx 0.50$ reported for similar-scale
cross-architecture pairs in prior work. We attribute this to three
factors. First, our validation set contains only 50 question-answer
pairs, while prior work uses 817 TruthfulQA questions for fitting;
larger fitting sets reliably improve $R^2$. Second, our models are
relatively small (1B and 1.5B parameters); prior work finds alignment
quality increases with model scale, with 8B-scale teachers achieving
the strongest results. Third, our Llama-Qwen pairing is cross-family;
intra-family pairings (e.g., Llama-8B $\to$ Llama-1B) achieve
$R^2 = 0.684$, substantially above cross-family pairs.
 
Importantly, none of these factors undermine the validity of our
end-to-end trained projections. Prior work reports a near-zero
correlation ($r \approx -0.07$) between offline geometric alignment
quality ($R^2$) and behavioral correction rate at test time,
concluding that directional accuracy in relevant subspaces matters
more than global variance explained. Our end-to-end training
procedure optimizes projection matrices directly against task
performance rather than $R^2$, and therefore implicitly optimizes
for precisely the directional accuracy that matters for downstream
behavior. The offline $R^2$ of 0.299 establishes that the
cross-architecture geometry is real and non-trivial; the end-to-end
training then refines the projection toward task-relevant directions
that an offline proxy dataset may not capture.
 
\section{OpenBookQA Training Curve}
\label{app:obqa_curve}
 
\begin{figure}[h]
\centering
\begin{tikzpicture}
\begin{axis}[
    width=0.82\columnwidth,
    height=5.5cm,
    xlabel={Training step},
    ylabel={Accuracy (\%)},
    xmin=0, xmax=1870,
    ymin=50, ymax=90,
    xtick={0,400,800,1240,1860},
    xticklabels={0,400,800,E2,E3},
    ytick={55,65,75,85},
    yticklabels={55\%,65\%,75\%,85\%},
    grid=major,
    grid style={line width=0.3pt, draw=gray!30},
    tick label style={font=\footnotesize},
    label style={font=\small},
    legend style={font=\footnotesize, at={(0.97,0.05)},
                  anchor=south east, draw=gray!40},
]
\addplot[color={rgb,1:red,0.357;green,0.310;blue,0.749},
         line width=1.2pt, mark=none]
coordinates {
    (30,74.0)(60,67.6)(90,75.0)(120,60.6)(150,76.8)
    (180,75.2)(210,78.6)(240,80.8)(270,79.0)(300,82.6)
    (330,83.2)(360,78.2)(390,79.4)(420,80.4)(450,81.0)
    (480,83.8)(510,82.4)(540,85.2)(570,85.2)(600,77.0)
    (630,83.0)(660,80.2)(690,82.0)(720,82.4)(750,82.6)
    (780,83.4)(840,84.6)(870,83.4)(900,82.6)(960,83.2)
    (990,85.4)(1050,83.2)(1110,84.4)(1140,85.0)(1260,83.4)
    (1290,84.2)(1320,84.8)(1350,85.2)(1440,84.4)(1560,84.8)
    (1650,84.8)(1770,85.0)(1800,84.4)(1830,84.0)
};
\addlegendentry{Val (mid-epoch)}
\addplot[color={rgb,1:red,0.753;green,0.314;blue,0.125},
         only marks, mark=*, mark size=2.5pt]
coordinates {
    (610, 77.6)
    (1230, 82.4)
    (1850, 82.8)
};
\addlegendentry{Test (end of epoch)}
\addplot[color=black!50, dashed, line width=0.9pt, mark=none]
coordinates {(0,76.6)(1870,76.6)};
\addlegendentry{Best single model (76.6\%)}
\addplot[color=gray!35, dotted, line width=0.7pt, mark=none]
    coordinates {(610,50)(610,90)};
\addplot[color=gray!35, dotted, line width=0.7pt, mark=none]
    coordinates {(1230,50)(1230,90)};
\end{axis}
\end{tikzpicture}
\caption{OpenBookQA validation accuracy over training. The network
crosses the best single-model baseline (dashed) during epoch~1 and
peaks at 85.4\% mid-epoch~2. Test accuracy at end of epoch~3 is
82.8\% (+6.2pp over best single model). Dotted vertical lines mark
epoch boundaries.}
\label{fig:obqa_curve}
\end{figure}
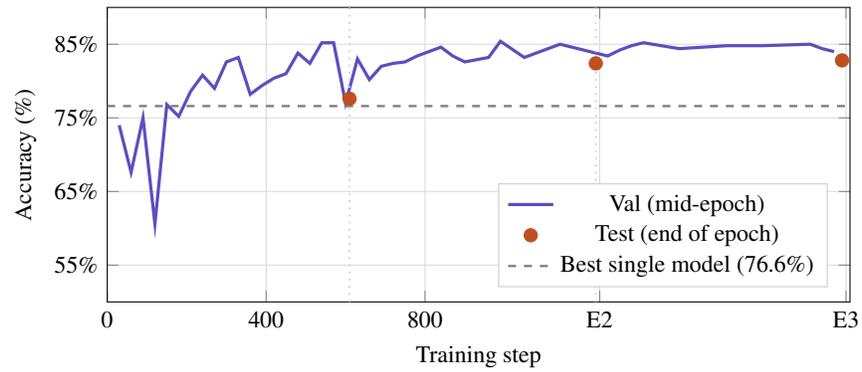

Figure~\ref{fig:obqa_curve} shows the validation accuracy trajectory
for the frozen LLM graph on OpenBookQA. The network reaches 74.0\%
validation accuracy within 30 gradient steps, crosses the best
single-model baseline (Qwen2.5-1.5B, 76.6\% test) before the end
of epoch~1, and peaks at 85.4\% mid-epoch~2. The curve plateaus
between epochs 2 and 3 (test accuracy 82.4\% $\to$ 82.8\%),
motivating our decision to stop training at epoch~3.
 
\end{document}